\documentclass{article}

\usepackage{arxiv}

\usepackage[utf8]{inputenc} 
\usepackage[T1]{fontenc}    
\usepackage{hyperref}       
\usepackage{url}            
\usepackage{booktabs}       
\usepackage{amsfonts}       
\usepackage{nicefrac}       
\usepackage{microtype}      
\usepackage{lipsum}		
\usepackage{graphicx}
\usepackage{natbib}
\usepackage{adjustbox}
\usepackage{float}
\usepackage{amsmath}
\usepackage{amssymb}
\usepackage{multirow}
\usepackage{multicol}
\usepackage{xcolor}
\usepackage{doi}

\title{TC-SSA: Token Compression via Semantic Slot Aggregation for Gigapixel Pathology Reasoning}

\date{} 					

\author{{\hspace{1mm}Zhuo Chen}\thanks{Zhuo Chen and Shawn Young contributed equally to this work and conducted it during an internship at Shenzhen University of Advanced Technology.} \\
	\textsuperscript{1}Shenzhen University of Advanced Technology\\
	Shenzhen, Guangdong, China \\
    \textsuperscript{2}University of Nottingham NingBo China\\
	NingBo, ZheJiang, China \\
    FoSE, Department of Mathematics with Applied Mathematics \\
	\texttt{smyzc9@nottingham.edu.cn} \\
	\And
	\hspace{1mm}Shawn Young\textsuperscript{*} \\
	\textsuperscript{1}Shenzhen University of Advanced Technology\\
	Shenzhen, Guangdong, China \\
    \And
	\hspace{1mm}Lijian Xu\thanks{Corresponding author.} \\
	Shenzhen University of Advanced Technology \\
	Shenzhen, Guangdong, China \\
	\texttt{xulijian@suat-sz.edu.cn} \\
}

\date{}

\renewcommand{\shorttitle}{TC-SSA for Gigapixel Spatial Reasoning}

\hypersetup{
pdftitle={TC-SSA: Token Compression via Semantic Slot Aggregation for Gigapixel Pathology Reasoning},
pdfsubject={cs.CV},
pdfauthor={Zhuo Chen, Shawn Young, Lijian Xu},
pdfkeywords={Gigapixel pathology, Token compression, Vision-language models},
}

\begin{document}

\maketitle

\begin{abstract}

The application of large vision-language models to computational pathology holds great promise for diagnostic assistants but faces a critical computational bottleneck: the gigapixel scale of Whole Slide Images (WSIs). A single WSI typically contains over $10^5$ patches, creating sequence lengths that exceed the constraints of standard Transformer architectures. Existing solutions often resort to spatial sampling, which risks discarding diagnostically critical evidence. 
To address this, we propose TC-SSA (\textbf{T}oken \textbf{C}ompression via \textbf{S}emantic \textbf{S}lot \textbf{A}ggregation), a learnable token compression framework that aggregates patch features into a fixed number of semantic slots.
A gated routing module assigns patches to slots using sparse Top-2 routing, followed by weighted aggregation, enabling global slide coverage under a strict token budget. The resulting representation retains diagnostically relevant information while reducing the number of visual tokens to 1.7\% of the original sequence.
On the SlideBench(TCGA), our model achieves 78.34\% overall accuracy and 77.14\% on the diagnosis subset, outperforming sampling-based baselines under comparable token budgets. The method also generalizes to MIL classification, reaching AUC of 95.83\% on TCGA-BRCA, 98.27\% on TCGA-NSCLC and 79.80\% on PANDA. These results suggest that learnable semantic aggregation provides an effective trade-off between efficiency and diagnostic performance for gigapixel pathology reasoning.
Our code is available at \url{https://anonymous.4open.science/r/TC-SSA/}.
\end{abstract}


\section{Introduction}

Recent progress in vision-language models (VLMs) \cite{yang2025learning,yang2025adapting,yu2025drift,yang2025resilient,xu2023learning} has enabled new forms of multimodal reasoning in computational pathology \citep{Lu2024VisualLanguage,yang2025walking}. These models can combine visual evidence with textual queries, offering a potential interface for slide-level question answering and clinical assistance. However, whole-slide images (WSIs) present a fundamental scalability challenge. A single WSI contains more than 
$10^5$ patches, and sequence lengths exceed the memory and computational limits of standard Transformer architectures. Direct end-to-end processing of all patches is therefore infeasible.

Although recent advancements in VLMs have enabled highly capable pathology copilots through diverse pretraining and unified architectures \citep{Huang2023VisualLanguage, Xu2024WholeSlide, Chen2024Towards, Lu2024VisualLanguage, Lu2024Multimodal, Sun2025CPathOmni, Zhang2023TextGuided, young2026scalar,yang2023t}, deploying these comprehensive models on gigapixel slides demands a critical balance between computational efficiency and diagnostic effectiveness. 
To address this challenge, existing approaches predominantly follow two directions. First, spatial sampling strategies, exemplified by LLaVA-Med \citep{Li2023LLaVAMed} and Quilt-LLaVA \citep{Seyfioglu2024QuiltLLaVA}, restrict the input to a fixed context window by discarding the majority of patches, which introduces a severe risk of omitting diagnostically critical regions. Second, sparse-attention frameworks such as SlideChat \citep{Chen2025SlideChat} retain broader visual evidence but incur substantially higher inference costs. Parallel to these primary directions, while foundational weakly supervised modeling is frequently formulated as Multiple Instance Learning \citep{Lu2021DataEfficient, Shao2021TransMIL, Hashimoto2024Multimodal, Wu2025Learning,yang2024segmentation}, recent investigations extensively explore token compression, instruction tuning, and summarization techniques to facilitate efficient slide-level learning for pathology models \citep{Tang2025Revisiting, Guo2025Focus, Chen2025CostEffective, Hu2025LoCPath, Wang2026WSISum, yang2025one, young2026fewer}.

To mitigate the efficiency bottleneck, we propose \textsc{TC-SSA} (Token Compression via Semantic Slot Aggregation), a learnable token-budgeting framework aggregating all patch features into fixed learnable semantic slots. To avoid systematically under-representing rare yet diagnostically critical evidence, a semantic affinity clustering objective further regularizes slot utilization. Under a fixed token budget, this bottleneck maintains global slide coverage, eliminating dense attention's quadratic cost.
Thus, our main contributions are:

1) \textbf{Semantic Slot-Based Token Compression:} We propose a semantic-driven mechanism for WSI token compression, routing all visual tokens into a fixed set of learned semantic slots based on shared contextual relevance instead of spatial proximity. It aggregates sparse, diagnostically critical evidence while suppressing redundant background noise, thereby preserving the global context of the image under a rigorous token budget.

2) \textbf{Robust Regularization for Semantic Slots:} We propose semantic affinity clustering, to mitigate slot collapse and ensure routing stability during training. By jointly optimizing a load-balancing loss, an entropy regularizer, and a z-loss, the framework strictly enforces balanced slot utilization and penalizes numerical instability. 

3) \textbf{Superior efficiency-performance tradeoff:} Under a strict visual-token budget (only 1.7\% of WSI), \textsc{TC-SSA} achieves 78.34\% overall accuracy on SlideBench(TCGA) and improves the Diagnosis subset to 77.14\%. Furthermore, it also demonstrates superior generalization with 55.94\% on  SlideBench(BCNB).

\section{Methodology}
\label{sec:methodology}

\subsection{Problem Formulation and Method Overview}

We consider a WSI as a massive collection of patches. Let the visual features of these patches, extracted by the pre-trained vision encoder, be denoted as an input sequence $X \in \mathbb{R}^{B\times N \times D}$. Here, $N$ represents the total number of patches, routinely exceeding $10^5$ for a single gigapixel slide, and $D$ denotes the feature dimension. The problem is defined as establishing a highly efficient compression function $f: \mathbb{R}^{B \times N \times D} \rightarrow \mathbb{R}^{B \times K \times D}$. This function projects the vast input sequence $X$ into a strictly bounded representation $X'$, where $K \ll N$ is a predefined visual token budget. The fundamental objective is achieving this extensive token reduction while meticulously preserving the original image's global semantic context, thereby fitting dense diagnostic evidence into the restricted context window of the VLM.

\begin{figure}[ht]
    \centering
    \includegraphics[width=0.99\textwidth]{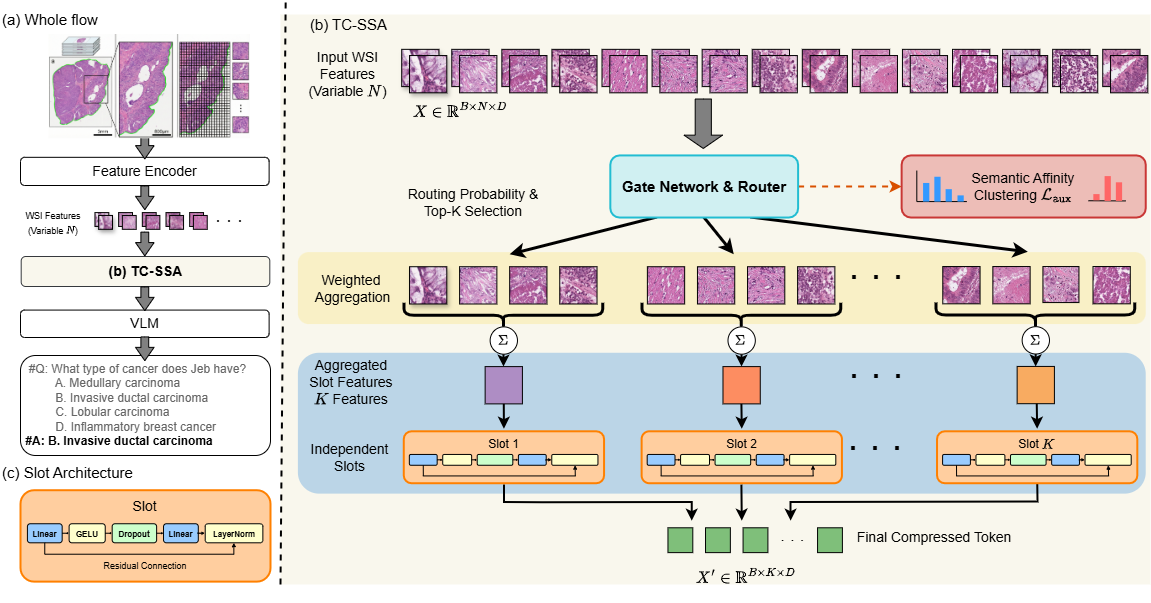}
    \caption{\textbf{Overview of the Token Compression via Semantic Slot Aggregation (TC-SSA) framework.} \textbf{(a) Whole flow:} a feature encoder extracts a variable-length feature sequence ($N$) from partitioned gigapixel WSI patches. The TC-SSA module compresses these dense features into a fixed token budget before feeding them into the VLM for downstream diagnostic reasoning. \textbf{(b) TC-SSA Module:} a gate network processes input features $X \in \mathbb{R}^{B \times N \times D}$ to determine routing probabilities, followed by Top-K selection. The auxiliary semantic affinity clustering objective ($\mathcal{L}_{aux}$) regularizes this routing distribution. Routed patch features undergo weighted aggregation into $K$ distinct slots, yielding the final compressed representation $X' \in \mathbb{R}^{B \times K \times D}$. \textbf{(c) Slot Architecture:} the internal configuration of each independent slot.
    }
    \label{fig:flowchart}
\end{figure}


To resolve this bottleneck, we introduce a learnable token-budgeting framework, Token Compression via Semantic Slot Aggregation (\textsc{TC-SSA}). As shown in Fig.\ref{fig:flowchart}, it comprises a gated routing mechanism and a slot-centric aggregation module. Initially, a lightweight gate computes a probability distribution over $K$ predefined semantic slots for each input patch. To enforce sparse assignment and manage computational costs, the system applies a Top-2 routing strategy, where each patch contributes to at most two optimal slots. Subsequently, routed patches are aggregated via weighted pooling to construct compact slot embeddings. Each aggregated slot is independently refined by a multilayer perceptron to produce the final compressed sequence $X'$. This sequence is then projected into the downstream VLM embedding space. To maintain routing stability and prevent slot collapse during training, the framework incorporates an auxiliary semantic affinity clustering objective.





\subsection{Semantic Slot Aggregation}
For each patch token $x_j$ within input sequence $X$, a lightweight gating module computes a probability distribution over $K$ distinct slots. This routing probability is $P(x_j) = \mathrm{Softmax}(W_g x_j)$, where $W_g \in \mathbb{R}^{B \times K \times D}$ represents the gate's learnable weight matrix. To enforce sparse assignment and control computational overhead, the system applies a Top-2 routing strategy. Rather than distributing patch information across all available slots, the framework isolates the two highest-probability slots. Let $m_{j,k} \in \{0, 1\}$ denote a binary mask indicating whether the $k$-th slot is among the top two selections for the $j$-th patch. Consequently, truncated routing weights are defined as $\tilde{P}_{j,k} = m_{j,k} P_{j,k}$. This hard truncation guarantees that every patch contributes to at most two semantic concepts, thereby stabilizing the routing process and strictly controlling the subsequent aggregation cost.

Following the determination of truncated routing probabilities, the framework executes a slot-centric feature aggregation step. This operation consolidates spatially dispersed patches sharing similar morphological or semantic properties into unified representations. For each semantic slot $k$, the aggregated feature vector $c_k$ is computed via a weighted pooling operation over all routed patches. The aggregation is formulated as 
\begin{equation}
c_k = 
\frac{
\sum_{j=1}^{N} \tilde{P}_{j,k} \, x_j
}{
\sum_{j=1}^{N} \tilde{P}_{j,k} + \delta
}
\end{equation}
where $\delta = 10^{-9}$ is a minor numerical constant preventing division by zero if a slot receives no assignments. By normalizing the sum of the features by the sum of the routed weights, the magnitude of the resulting token remains stable regardless of the number of patches assigned to the specific slot. This mechanism effectively maps the high-dimensional visual evidence from the slide's physical coordinate space into a highly compressed, fixed-budget semantic space \citep{young2026scalar}. Through this procedure, critical diagnostic cues physically scattered across the tissue are successfully distilled into compact visual tokens without sacrificing global context.




\subsection{Robust Regularization for Semantic Slots}
To prevent slot collapse, where a disproportionate number of patches are routed to a single semantic concept, the framework incorporates an auxiliary semantic affinity clustering objective during training. Without proper regularization, the gating module tends to heavily favor a few specific slots, which severely limits the representational capacity of the compressed sequence. To counteract this phenomenon, the primary component of the auxiliary objective is a load-balancing loss, denoted as 
\begin{equation}
\mathcal{L}_{\text{switch}}
= K \sum_{k=1}^{K} P_k f_k
\end{equation}
where $P_k$ is the average routing probability to slot $k$ and $f_k$ is the fraction of patches assigned to slot $k$ under Top-2 routing. This loss formulation, widely adopted in sparse routing architectures, penalizes the network if the fraction of patches routed to a given slot deviates significantly from a uniform distribution. By enforcing a balanced utilization of all $K$ available slots, the load-balancing loss guarantees the diverse morphological patterns within the gigapixel image are adequately captured across the entire semantic budget.

Furthermore, the auxiliary objective integrates two additional regularization terms to enhance routing stability. An entropy regularizer is applied to the gate's probability distribution to discourage overly confident but incorrect routing decisions early in training, denoted as:
\begin{equation}
\mathcal{L}_{\text{ent}}
= 1 - 
\frac{
- \sum_{k=1}^{K} P_k \log (P_k + \epsilon)
}{
\log K
}
\end{equation}
with $\epsilon = 10^{-8}$ for numerical stability. Additionally, a z-loss term $\mathcal{L}_{z}$ penalizes excessively large logit magnitudes produced by the gating network, thereby preventing numerical instability:
\begin{equation}
\mathcal{L}_{z} = \alpha \frac{1}{N} \sum_{j=1}^{N} \left( \log \sum_{k=1}^{K} \exp(g_{j,k}) \right)^2
\end{equation}
where $g_{j,k}$ denotes the pre-softmax gate logits and $\alpha = 10^{-4}$. The auxiliary routing regularization is optimized jointly with the primary task loss, denoted as $\mathcal{L}_{task}$, of the downstream VLM. Specifically, the overall training objective is formulated as:
\begin{equation}
    \mathcal{L}_{total} = \mathcal{L}_{task} + \lambda (\mathcal{L}_{\text{switch}} + 0.5\,\mathcal{L}_{\text{ent}} + \mathcal{L}_{z})
\end{equation}
 where $\lambda$ is a hyperparameter balancing the contribution of these auxiliary terms. This regularization ensures diverse and stable semantic slots, yielding compact and informative patch representations.

\section{Experiments and Results}
\label{sec:experiments}

Our model is trained on SlideBench following \citep{Chen2025SlideChat}, while using 2$\times$A6000 GPUs instead of 8$\times$A100 of \citep{Chen2025SlideChat}.
The benchmark defines \textbf{Microscopy} as low-level morphology and staining, and \textbf{Diagnosis} as clinical reasoning for grading and subtyping. \textbf{Overall Accuracy} is computed across all multiple-choice VQA pairs. We use a frozen \textsc{CONCH} encoder~\citep{Lu2024VisualLanguage} to extract patch features. The model uses $K=32$ slots and auxiliary routing loss weight $\lambda = 0.1$ with Top-2 routing.
We performed the comparative studies and ablation studies as following.

\begin{table}[H]
\centering
\caption{Comparison of performance on the SlideBench(TCGA), and zero-shot results on SlideBench(BCNB) and WSI-VQA*. 
SlideChat\cite{Chen2025SlideChat} is treated as the upper-bound for full uncompressed WSI inference. 
Flops and accuracy are used as the metric.
Best results are highlighted in red.}
\label{tab:combined_results}
\begin{adjustbox}{width=.85\textwidth}{
\begin{tabular}{lcccccc}
\toprule
\multirow{2}{*}{\textbf{Methods}} & \multirow{2}{*}{\textbf{Flops}} & \multicolumn{3}{c}{\textbf{SlideBench(TCGA)}} & \textbf{SlideBench} & \textbf{WSI} \\
\cmidrule(lr){3-5} 
 & & \textbf{Microscopy} & \textbf{Diagnosis} & \textbf{Overall} & \textbf{(BCNB)} & \textbf{(VQA*)} \\
\midrule
\multicolumn{7}{c}{\textcolor{gray}{\textbf{(Upper Bound)}}} \\
\textcolor{gray}{SlideChat \cite{Chen2025SlideChat}} & \textcolor{gray}{133.3T} & \textcolor{gray}{87.64} & \textcolor{gray}{73.27} & \textcolor{gray}{81.17} & \textcolor{gray}{54.14} & \textcolor{gray}{60.18} \\
\midrule
\multicolumn{7}{c}{\textbf{(Token Compression $\sim60\times$)}} \\
Random Baseline & & 24.44 & 24.91 & 25.02 & 24.40 & 24.14 \\
GPT-4o \cite{OpenAI2024GPTo} & & 62.89 & 46.69 & 57.91 & 41.43 & 30.41 \\
LLaVA-Med \cite{Li2023LLaVAMed} & 1.70T & 47.34 & 32.78 & 42.00 & 30.10 & 26.31 \\
Quilt-LLaVA \cite{Seyfioglu2024QuiltLLaVA} & 1.70T & 57.76 & 35.96 & 48.07 & 32.19 & 44.43 \\
MedDr \cite{He2024MedDr} & 1.70T & 73.30 & 57.78 & 67.70 & 33.67 & 54.36 \\
\midrule
\textbf{TC-SSA (Ours)} & 1.72T & \textcolor{red}{\textbf{81.94}} & \textcolor{red}{\textbf{77.14}} & \textcolor{red}{\textbf{78.34}} & \textcolor{red}{\textbf{55.94}} & \textcolor{red}{\textbf{56.62}} \\
\bottomrule
\end{tabular}
}\end{adjustbox}
\end{table}


\noindent\textbf{1) Efficiency-Performance Comparative Studies.}
As shown in Table~\ref{tab:combined_results}, \textsc{TC-SSA} achieves the best overall accuracy of \textbf{78.34\%} on SlideBench(TCGA). 
Despite operating under the same token budget, \textsc{TC-SSA} surpasses SlideChat on the Diagnosis subset, highlighting the effectiveness of the routing-and-pooling strategy, which removes redundant patches while preserving critical evidence.
Moreover, the superiority of \textsc{TC-SSA} further generalises to zero-shot settings, where it consistently outperforms competitors on both SlideBench (BCNB) and WSI-VQA*, demonstrating strong generalization of our method.
\begin{figure}[H]
    \centering
    \includegraphics[width=0.85\linewidth]{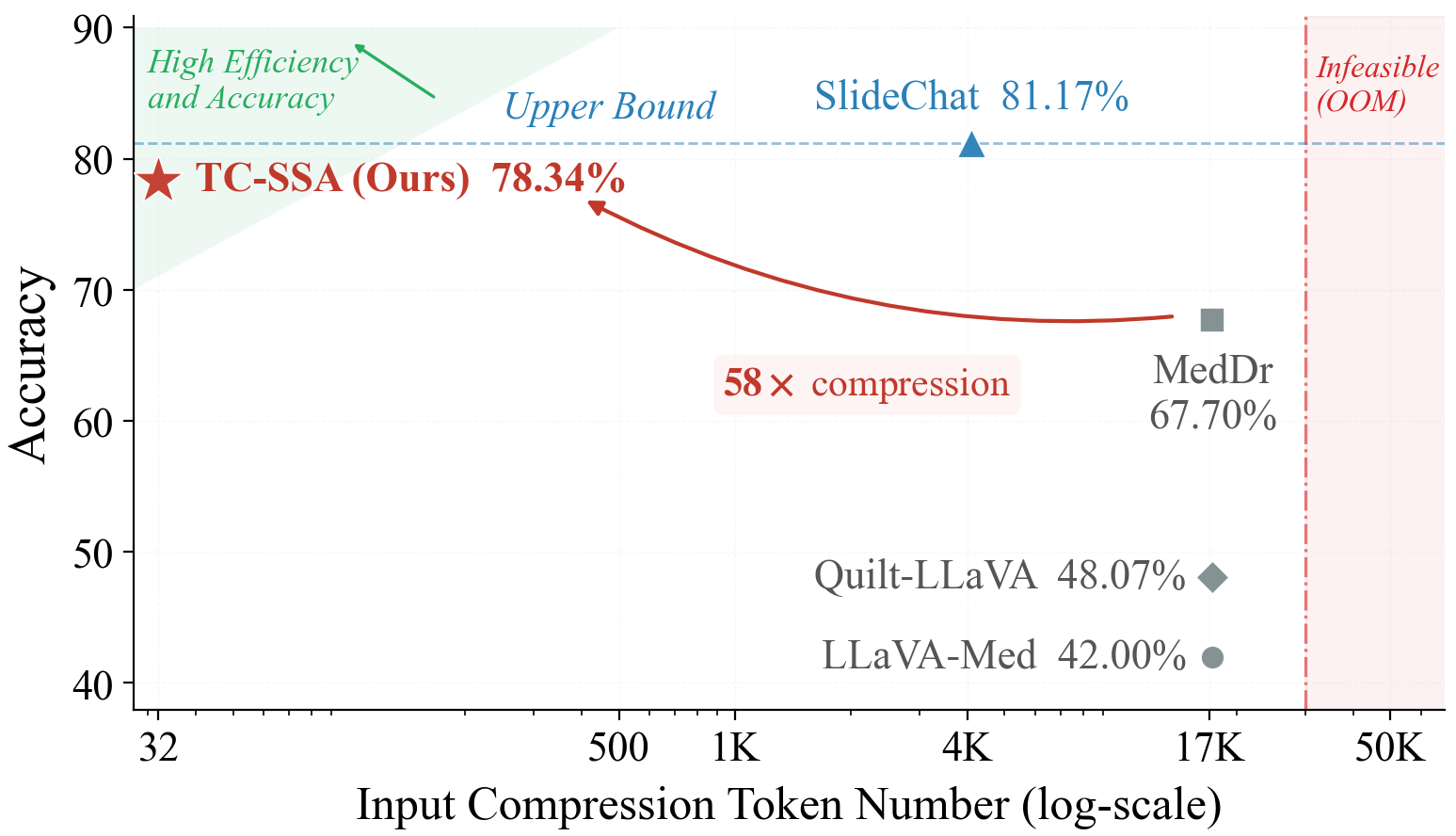}
    \caption{
    Using only 32 visual tokens, TC-SSA achieves 78.34\% overall accuracy on SlideBench(TCGA) benchmark, yielding a 58$\times$ compression ratio compared to original patch features. SlideChat is reported as an uncompressed reference, while full-WSI inference is infeasible due to out-of-memory (OOM) constraints.
    }
    \label{fig:optimize}
\end{figure}

\begin{figure}[H]
    \centering
    \includegraphics[width=0.85\linewidth]{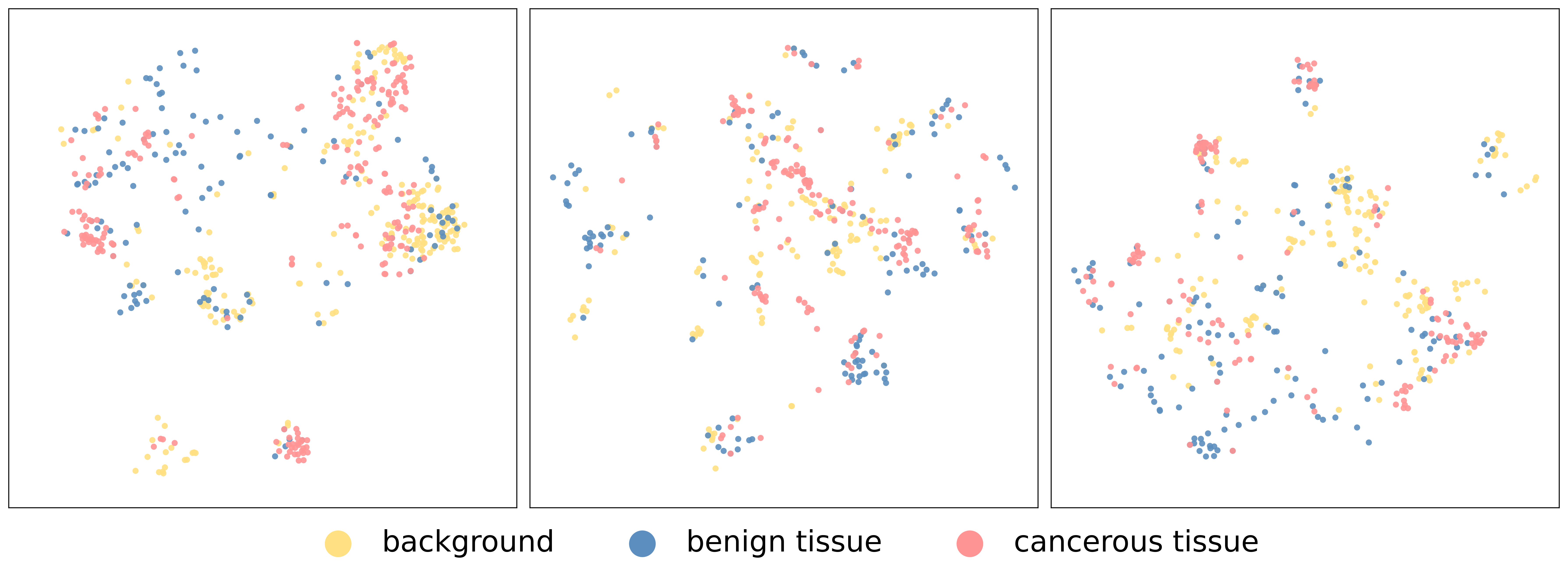}
    \caption{Multi-dataset t-SNE of patch embeddings grouped by slot. Colored clusters correspond to different tissue semantics, showing slot assignment induces coherent, dataset-consistent grouping in embedding space.}
    \label{fig:tsne}
\end{figure}

\textbf{Figure \ref{fig:optimize}} demonstrates superior performance on SlideBench(TCGA) of the proposed method. SlideChat serves as an upper bound without an explicit token compression, whereas full-resolution WSI inference is impractical due to OOM constraints. By condensing global evidence into $K$ tokens rather than discarding patches (e.g., LLaVA-Med, Quilt-LLaVA), our method outperforms sampling baselines by 10.64\% in accuracy with 98.3\% fewer input tokens. 
Its linear-time complexity $O(N\cdot K)$ enables practical deployment in memory and latency-constrained clinical settings. We further analyze the semantic patterns of each slot across multiple TCGA cohorts. \textbf{Figure~\ref{fig:tsne}} shows patches routed to the same slot cluster together in embedding space across datasets, suggesting the gate consistently groups similar tissue patterns.

\newpage

\noindent\textbf{2) Ablation Studies.} 
\begin{figure}[H]
\noindent
\centering
\includegraphics[width=0.8\linewidth, height=0.275\textheight]{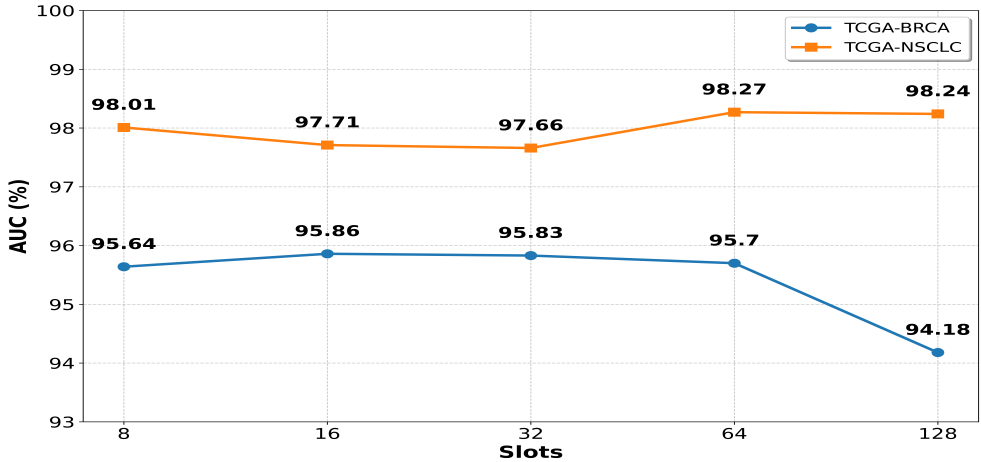}
\caption{Ablation experiments of the number of semantic slots $K$ on TCGA-BRCA and TCGA-NSCLC. We set $K=32$ by default.}
\label{fig:auc_slots}
\end{figure}
We firstly performed the ablation experiments of choosing the number of slots.
Figure~\ref{fig:auc_slots} reports MIL AUC as a function of the slot budget $K$. For TCGA-BRCA, performance is stable for $K\in\{16,32,64\}$ and drops at overly large budgets (e.g., $K=128$), suggesting diminishing returns and possible over-fragmentation of semantic evidence. For TCGA-NSCLC, larger $K$ brings only marginal gains beyond $K=32$.
Employing Top-2 assignment with an auxiliary routing loss weight $\lambda = 0.1$ was critical for stable clustering. Reducing $\lambda$ resulted in slot collapse, where a single slot absorbed over 60\% of patches, significantly degrading performance on heterogeneous cases. Increasing $K$ caused slots to aggregate task-irrelevant artifacts and background noise, thereby diluting the discriminative signals passed to the final classifier.


We further conducted an ablation study by replacing the default CONCH encoder\citep{Lu2024VisualLanguage} with the UNI encoder\citep{Chen2024Towards}. Notably, when equipped with the same UNI encoder as recent baselines, our framework still achieves state-of-the-art (SOTA) performance. This demonstrates that our approach is highly generalizable and can yield highly competitive results even when utilizing an encoder different from CONCH.

\begin{table}[ht]
\centering
\caption{Ablation experiments of the encoder. AUC scores of typical methods on TCGA and PANDA MIL benchmarks. Best results are highlighted in red.}
\label{tab:tcga_mil_auc}
\begin{adjustbox}{width=.80\textwidth}{
    \begin{tabular}{lccc}
    \toprule
    \textbf{Method} & \textbf{TCGA-BRCA} & \textbf{TCGA-NSCLC} &\textbf{PANDA}\\
    \midrule
    ABMIL\citep{Ilse2018Attentionbased} & $94.05\%$ & $97.04\%$ & 74.69\%\\
    TransMIL\citep{Shao2021TransMIL} & $93.33\%$ & $97.27\%$ & 68.06\%\\
    RRTMIL\citep{Tang2024Feature} & $94.61\%$ & $97.88\%$ & 74.93\%\\
    2DMamba\citep{Zhang2024DMamba} & $93.08\%$ & $97.14\%$ & 76.37\%\\
    TC-SSA (Ours) & \textcolor{red}{\textbf{95.83\%}} & \textcolor{red}{\textbf{98.27\%}} & \textcolor{red}{\textbf{79.80\%}}\\
    \bottomrule
    \end{tabular}
}\end{adjustbox}
\end{table}



\section{Conclusion}
\label{sec:conclusion}

We presented \textsc{TC-SSA}, a token compression module for gigapixel pathology VLMs utilizing semantic slot aggregation. Under a strict visual-token budget, \textsc{TC-SSA} achieves 78.34\% overall accuracy on SlideBench(TCGA), improving the \textit{Diagnosis} subset to 77.14\%, while attaining AUC of 95.83\% on TCGA-BRCA and 98.27\% on TCGA-NSCLC. This demonstrates learnable aggregation's strong performance under constrained budgets. Currently, a fixed slide-level slot budget $K$ makes compression quality dependent on the patch encoder, and trading fine-grained spatial geometry for semantic structure may affect localization-heavy tasks.

\newpage
\renewcommand{\bibname}{references}
\bibliographystyle{unsrtnat}
\bibliography{references}

@inproceedings{Tang2024Feature,
  author = {Tang, Wenhao and Zhou, Fengtao and Huang, Shengyue and Zhu, Xiang and Zhang, Yi and Liu, Bo},
  title = {Feature Re-Embedding: Towards Foundation Model-Level Performance in Computational Pathology},
  booktitle = {CVPR},
  year = {2024},
  eprint = {2402.17228},
  archiveprefix = {arXiv}
}

@inproceedings{Ilse2018Attentionbased,
  author = {Ilse, Maximilian and Tomczak, J. and Welling, M.},
  title = {Attention-based Deep Multiple Instance Learning},
  booktitle = {ICML},
  year = {2018},
  eprint = {1802.04712},
  archiveprefix = {arXiv}
}

@inproceedings{Zhang2024DMamba,
  author = {Zhang, Jingwei and Nguyen, Anh Tien and Han, Xi and Trinh, Vincent Quoc-Huy and Qin, Hong and Samaras, Dimitris and Hosseini, Mahdi S.},
  title = {2DMamba: Efficient State Space Model for Image Representation with Applications on Giga-Pixel Whole Slide Image Classification},
  booktitle = {CVPR},
  year = {2024},
  eprint = {2412.00678},
  archiveprefix = {arXiv}
}

@inproceedings{Wu2025Learning,
  author = {Wu, Junxian and Chen, Minheng and Ke, Xinyi and Xun, Tianwang and Jiang, Xiaoming and Zhou, Hongyu and Shao, Lizhi and Kong, Youyong},
  title = {Learning Heterogeneous Tissues with Mixture of Experts for Gigapixel Whole Slide Images},
  booktitle = {CVPR},
  year = {2025}
}

@inproceedings{Shao2021TransMIL,
  author = {Shao, Zhuchen and Bian, Hao and Chen, Yang and Wang, Yifeng and Zhang, Jian and Ji, Xiangyang and Zhang, Yongbing},
  title = {TransMIL: {Transformer} based Correlated Multiple Instance Learning for Whole Slide Image Classification},
  booktitle = {NeurIPS},
  year = {2021}
}

@article{He2024MedDr,
  author = {He, Sunan and Nie, Yuxiang and Chen, Zhixuan and Cai, Zhiyuan and Wang, Hongmei and Yang, Shu and Chen, Hao},
  title = {MedDr: Diagnosis-Guided Bootstrapping for Large-Scale Medical Vision-Language Learning},
  journal = {CoRR},
  year = {2024},
  eprint = {2404.15127},
  archiveprefix = {arXiv}
}

@inproceedings{Li2023LLaVAMed,
  author = {Li, Chunyuan and Wong, Cliff and Zhang, Sheng and Usuyama, Naoto and Liu, Haotian and Yang, Jianwei and Naumann, Tristan and Poon, Hoifung and Gao, Jianfeng},
  title = {LLaVA-Med: Training a Large Language-and-Vision Assistant for Biomedicine in One Day},
  booktitle = {NeurIPS},
  year = {2023}
}

@inproceedings{Seyfioglu2024QuiltLLaVA,
  author = {Seyfioglu, Mehmet Saygin and Ikezogwo, Wisdom Oluchi and Ghezloo, Fatemeh and Krishna, Ranjay and Shapiro, Linda G.},
  title = {Quilt-LLaVA: Visual Instruction Tuning by Extracting Localized Narratives from Open-Source Histopathology Videos},
  booktitle = {CVPR},
  year = {2024}
}

@article{Hashimoto2024Multimodal,
  author = {Hashimoto, Noriaki and Hanada, Hiroyuki and Miyoshi, Hiroaki and Nagaishi, Miharu and Sato, Kensaku and Hontani, Hidekata and Ohshima, Koichi and Takeuchi, Ichiro},
  title = {Multimodal Gated Mixture of Experts Using Whole Slide Image and Flow Cytometry for Multiple Instance Learning Classification of Lymphoma},
  journal = {Journal of Pathology Informatics},
  year = {2024}
}

@inproceedings{Sun2025CPathOmni,
  author = {Sun, Yuxuan and Si, Yixuan and Zhu, Chenglu and Gong, Xuan and Zhang, Kai and Chen, Pingyi and Zhang, Ye and Shui, Zhongyi and Lin, Tao and Yang, Lin},
  title = {CPath-Omni: {A} Unified Multimodal Foundation Model for Patch and Whole Slide Image Analysis in Computational Pathology},
  booktitle = {CVPR},
  year = {2025}
}

@article{Xu2024WholeSlide,
  author = {Xu, Hanwen and Usuyama, Naoto and Bagga, Jaspreet and Zhang, Sheng and Rao, Rajesh and Naumann, Tristan and Wong, Cliff and Gero, Zelalem and González, Javier and Gu, Yu and others},
  title = {A Whole-Slide Foundation Model for Digital Pathology from Real-World Data},
  journal = {Nature},
  year = {2024}
}

@article{Lu2024Multimodal,
  author = {Lu, Ming Y and Chen, Bowen and Williamson, Drew F K and Chen, Richard J and Zhao, Melissa and Chow, Aaron K and Ikemura, Kenji and Kim, Ahrong and Pouli, Dimitra and Patel, Ankush and others},
  title = {A Multimodal Generative AI Copilot for Human Pathology},
  journal = {Nature},
  year = {2024}
}

@article{Lu2024VisualLanguage,
  author = {Lu, Ming Y and Chen, Bowen and Williamson, Drew F K and Chen, Richard J and Liang, Ivy and Ding, Tong and Jaume, Guillaume and Odintsov, Igor and Le, Long Phi and Gerber, Georg and others},
  title = {A Visual-Language Foundation Model for Computational Pathology},
  journal = {Nature medicine},
  year = {2024}
}

@article{Chen2024Towards,
  author = {Chen, Richard J and Ding, Tong and Lu, Ming Y and Williamson, Drew F K and Jaume, Guillaume and Song, Andrew H and Chen, Bowen and Zhang, Andrew and Shao, Daniel and others},
  title = {Towards a general-purpose foundation model for computational pathology},
  journal = {Nature medicine},
  year = {2024}
}

@article{Huang2023VisualLanguage,
  author = {Huang, Zhi and Bianchi, Federico and Yuksekgonul, Mert and Montine, Thomas J and Zou, James},
  title = {A Visual-Language Foundation Model for Pathology Image Analysis Using Medical Twitter},
  journal = {Nature medicine},
  year = {2023}
}

@inproceedings{Guo2025Focus,
  author = {Guo, Zhengrui and Xiong, Conghao and Ma, Jiabo and Sun, Qichen and Feng, Lishuang and Wang, Jinzhuo and Chen, Hao},
  title = {Focus: Knowledge-Enhanced Adaptive Visual Compression for Few-Shot Whole Slide Image Classification},
  booktitle = {CVPR},
  year = {2025}
}

@article{Lu2021DataEfficient,
  author = {Lu, Ming Y and Williamson, Drew F K and Chen, Tiffany Y and Chen, Richard J and Barbieri, Matteo and Mahmood, Faisal},
  title = {Data-Efficient and Weakly Supervised Computational Pathology on Whole-Slide Images},
  journal = {Nature biomedical engineering},
  year = {2021}
}

@inproceedings{Chen2025SlideChat,
  author = {Chen, Ying and Wang, Guoan and Ji, Yuanfeng and Li, Yanjun and Ye, Jin and Li, Tianbin and Hu, Ming and Yu, Rongshan and Qiao, Yu and He, Junjun},
  title = {SlideChat: A Large Vision-Language Assistant for Whole-Slide Pathology Image Understanding},
  journal = {arXiv preprint arXiv:2410.11761},
  booktitle = {CVPR},
  year = {2025},
  eprint = {2410.11761},
  archiveprefix = {arXiv}
}

@inproceedings{Tang2025Revisiting,
  author = {Tang, Wenhao and Qin, Rong and Fang, Heng and Zhou, Fengtao and Chen, Hao and Li, Xiang and Cheng, Ming-Ming},
  title = {Revisiting End-to-End Learning with Slide-Level Supervision in Computational Pathology},
  journal = {arXiv preprint arXiv:2506.02408},
  booktitle = {NeurIPS},
  year = {2025}
}

@article{OpenAI2024GPTo,
  author = {OpenAI},
  title = {{GPT}-4o System Card},
  journal = {arXiv preprint arXiv:2410.21276},
  year = {2024},
  eprint = {2410.21276},
  archiveprefix = {arXiv}
}

@article{Hu2025LoCPath,
  author = {Hu, Qingqiao and Lyu, Weimin and Xu, Meilong and Qi, Kehan and Hu, Xiaoling and Gupta, Saumya and Zhou, Jiawei and Chen, Chao},
  title = {LoC-Path: Learning to Compress for Pathology Multimodal Large Language Models},
  journal = {arXiv preprint arXiv:2512.05391},
  year = {2025},
  eprint = {2512.05391},
  archiveprefix = {arXiv}
}

@inproceedings{Zhang2023TextGuided,
  author = {Zhang, Yunkun and Gao, Jin and Zhou, Mu and Wang, Xiaosong and Qiao, Yu and Zhang, Shaoting and Wang, Dequan},
  title = {Text-Guided Foundation Model Adaptation for Pathological Image Classification},
  booktitle = {MICCAI},
  year = {2023}
}

@article{Chen2025CostEffective,
  author = {Chen, Kaitao and Liu, Mianxin and Yan, Fang and Ma, Lei and Shi, Xiaoming and Wang, Lilong and Wang, Xiaosong and Zhu, Lifeng and Wang, Zhe and Zhou, Mu and others},
  title = {Cost-Effective Instruction Learning for Pathology Vision and Language Analysis},
  journal = {Nature Computational Science},
  year = {2025}
}

@article{Wang2026WSISum,
  author = {Wang, Baizhi and Zhang, Kun and Wang, Yuhao and Gu, Yunjie and Luan, Haijing and Zhou, Ying and Hu, Taiyuan and Wang, Rundong and Yang, Zhidong and Jiang, Zihang and others},
  title = {WSISum: WSI Summarization via Dual-Level Semantic Reconstruction},
  journal = {Med. Image Anal.},
  year = {2026}
}

@inproceedings{yang2025one,
  title={One Leaf Reveals the Season: Occlusion-Based Contrastive Learning with Semantic-Aware Views for Efficient Visual Representation},
  author={Yang, Xiaoyu and Xu, Lijian and Li, Hongsheng and Zhang, Shaoting},
  booktitle={International Conference on Machine Learning},
  pages={71425--71440},
  year={2025},
  organization={PMLR}
}

@article{young2026fewer,
  title={Fewer Tokens, Greater Scaling: Self-Adaptive Visual Bases for Efficient and Expansive Representation Learning},
  author={Young, Shawn and Zeng, Xingyu and Xu, Lijian},
  journal={arXiv preprint arXiv:2511.19515},
  year={2025}
}

@article{young2026scalar,
  title={SCALAR: Spatial-Concept Alignment for Robust Vision in Harsh Open World},
  author={Yang, Xiaoyu and Xu, Lijian and Zeng, Xingyu and Wang, Xiaosong and Li, Hongsheng and Zhang, Shaoting},
  journal={Pattern Recognition},
  pages={113203},
  year={2026},
  publisher={Elsevier}
}

@article{xu2023learning,
  title={Learning a multi-task transformer via unified and customized instruction tuning for chest radiograph interpretation},
  author={Xu, Lijian and Ni, Ziyu and Liu, Xinglong and Wang, Xiaosong and Li, Hongsheng and Zhang, Shaoting},
  journal={arXiv preprint arXiv:2311.01092},
  year={2023}
}

@article{yang2024segmentation,
  title={Segmentation and vascular vectorization for coronary artery by geometry-based cascaded neural network},
  author={Yang, Xiaoyu and Xu, Lijian and Yu, Simon and Xia, Qing and Li, Hongsheng and Zhang, Shaoting},
  journal={IEEE Transactions on Medical Imaging},
  volume={44},
  number={1},
  pages={259--269},
  year={2024},
  publisher={IEEE}
}

@article{yang2025learning,
  title={Learning from All: Concept Alignment for Autonomous Distillation from Multiple Drifting MLLMs},
  author={Yang, Xiaoyu and Lu, Jie and Yu, En},
  journal={arXiv preprint arXiv:2510.04142},
  year={2025}
}

@inproceedings{yang2025adapting,
  title={Adapting Multi-modal Large Language Model to Concept Drift From Pre-training Onwards},
  author={Xiaoyu Yang and Jie Lu and En Yu},
  booktitle={The Thirteenth International Conference on Learning Representations},
  year={2025},
}

@article{yu2025drift,
  title={Drift-aware collaborative assistance mixture of experts for heterogeneous multistream learning},
  author={Yu, En and Lu, Jie and Wang, Kun and Yang, Xiaoyu and Zhang, Guangquan},
  journal={arXiv preprint arXiv:2508.01598},
  year={2025}
}

@inproceedings{yang2025walking,
  title={Walking the Tightrope: Autonomous Disentangling Beneficial and Detrimental Drifts in Non-Stationary Custom-Tuning},
  author={Yang, Xiaoyu and Lu, Jie and Yu, En},
  booktitle={The Thirty-ninth Annual Conference on Neural Information Processing Systems},
  year={2025}
}

@inproceedings{yang2023t,
  title={T-distributed spherical feature representation for imbalanced classification},
  author={Yang, Xiaoyu and Chen, Yufei and Yue, Xiaodong and Xu, Shaoxun and Ma, Chao},
  booktitle={Proceedings of the AAAI Conference on Artificial Intelligence},
  volume={37},
  number={9},
  pages={10825--10833},
  year={2023}
}

@article{yang2025resilient,
  title={Resilient Contrastive Pre-training under Non-Stationary Drift},
  author={Yang, Xiaoyu and Lu, Jie and Yu, En and Duan, Wei},
  journal={arXiv preprint arXiv:2502.07620},
  year={2025}
}

\end{document}